\documentclass[letterpaper, 10 pt, conference]{ieeeconf}  
\usepackage[spanish]{babel}
\usepackage[utf8]{inputenc}
\usepackage{listings}
\usepackage{caption}
\usepackage{DejaVuSans}
\usepackage{DejaVuSansMono}
\usepackage{amssymb}
\usepackage{amsmath}

\usepackage{subcaption}
\usepackage{graphicx}

\usepackage{alltt}

\usepackage{fancyvrb}

\usepackage[table,xcdraw]{xcolor}
\usepackage{multirow}

\usepackage[T1]{fontenc}
\usepackage{fancyvrb}
\usepackage{listings}
\usepackage[scaled=.55]{beramono}    
\fvset{baselinestretch=0.5}
\DeclareCaptionFormat{listing}{\rule{\dimexpr0.9\columnwidth+17pt\relax}{0.35pt}\par\vskip1pt#1#2#3}
\newsavebox{\FVerbBox}
\usepackage[most]{tcolorbox}
\tcbset{
    frame code={}
    center title,
    left=0pt,
    right=0pt,
    top=0pt,
    bottom=0pt,
    colback=gray!10,
    colframe=white,
    width=0.45\textwidth,
    enlarge left by=0mm,
    boxsep=5pt,
    arc=0pt,outer arc=0pt,
    }

\IEEEoverridecommandlockouts                              

\title{\LARGE \bf
Time-Sensitive Networking for robotics
}
\author{Carlos San Vicente Gutiérrez, Lander Usategui San Juan, Irati Zamalloa Ugarte, Víctor Mayoral Vilches \thanks{Erle Robotics} %
}

\usepackage{blindtext}
\usepackage{graphicx}
\usepackage{listings}
\usepackage{xcolor}

\usepackage{hyperref}
\hypersetup{
    colorlinks=true,
    linkcolor=blue,
    filecolor=magenta,      
    urlcolor=cyan,
    citecolor=blue,
}

\usepackage[normalem]{ulem}

\begin{document}
\maketitle

\begin{abstract}

We argue that Time-Sensitive Networking (TSN) will become the de facto standard for real-time communications in robotics. We present a review and  classification of the different communication standards which are relevant for the field and introduce the typical problems with traditional switched Ethernet networks. We discuss some of the TSN features relevant for deterministic communications and evaluate experimentally one of the shaping mechanisms –the time-aware shaper– in an exemplary robotic scenario. In particular, and based on our results, we claim that many of the existing real-time industrial solutions will slowly be replaced by TSN. And that this will lead towards a unified landscape of physically interoperable robot and robot components.



\end{abstract}

\section{Introduction}
\label{introduction}


%

The field of robotics is growing rapidly. New areas such as professional, consumer or industrial robotics are demanding more flexible technologies and a set of standardized policies that facilitate the process of designing, manufacturing and configuring a robot for potentially more than one specific application. Previous work \cite{8046383, 2018arXiv180201459Z} highlighted the relevance of standard interfaces at different levels.\\

One of the main problems in robotics, as it happens in other industries, is that there is no such thing as a standard communication protocol, but a variety of them. Choosing a communication protocol is not straightforward: the list is large, and each protocol has evolved to meet the needs of a particular application area. Typically, each protocol has been customized for specific applications and, as a result, multiple communication protocols and buses are used to meet different requirements within those more complex use cases. In fact, many industrial protocols have common technological baselines, but customize upper abstraction layers to meet different requirements as pictured in Figure \ref{stack} for real-time Ethernet solutions. In the case of  real-time communication protocols, the links and physical layers are commonly modified to achieve a better performance. This leads to hardware incompatibility problems, making communications between devices cumbersome. A common solution is the use of gateways (or bridges), which add cost, complexity and produce a loss in performance. Having a unique standard protocol would improve the interoperability between robots and facilitate the robotic component integration, which is still one of the main hurdles in the robot building process, as stated at \cite{2018arXiv180204082M}. Robotic peripheral manufacturers suffer especially from these problems because they need to support several protocols, further increasing the integration time and costs. \\

Time-Sensitive Networking (TSN) is a set of standards defined by the Time-Sensitive Networking task group of the IEEE 802.1 working group designed to make Ethernet more deterministic. The TSN sub-standards were created to meet different communication requirements in the industry: automation, automotive, audio, video, etc.

\begin{figure}[h!]
\centering
 \includegraphics[width=0.4\textwidth]{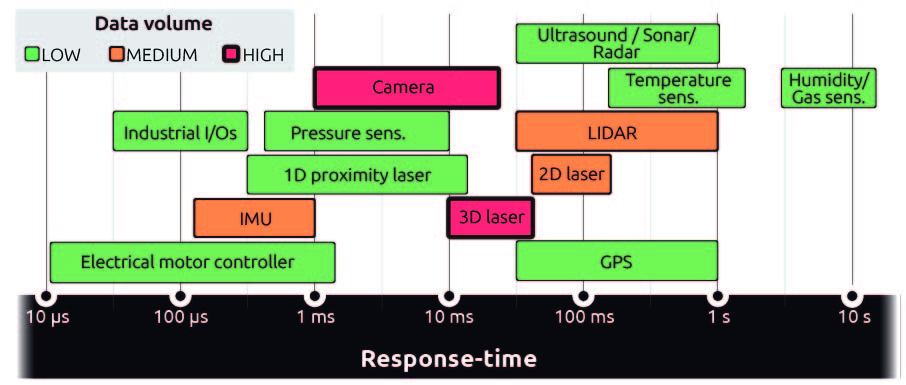}
\caption{\footnotesize Typical response-time of common robotic components. For sensors, the response-time reflects the typical time required to provide digital data of their measurements. For actuators, it states the typical control cycle.}
\label{response}
\end{figure}

Most of the existing real-time Ethernet solutions were created for low data volume applications such as distributed motion control. These solutions are usually very limited in bandwidth and cannot reach the Ethernet bandwidth capabilities. With the growing integration in robotics of Artificial Intelligence (AI), computer vision or predictive maintenance to name a few, there is an increasing need of sensors and actuators streaming high bandwidth data in real-time. The information provided by these sensors is often integrated in the control system or needs to be monitored in real-time. The common solution is to use a specific bus for real-time control and a separate one for higher bandwidth communications. As more and more high bandwidth traffic is generated, the control process of having two separated communications is inefficient. Figure \ref{response} shows the diversity of several robot components and their different networking requirements. Motors usually require simple data as parameters, such as set-points like position, velocity, torque; a camera system streams instead a considerably larger amount of data, which can go up to few megabytes per second. Adding real-time capabilities to Ethernet, TSN provides a common communication channel for high bandwidth traffic and real-time control traffic.



TSN will also improve the access to the robot components which is especially interesting for predictive maintenance, re-configurability or adaptability\cite{2018arXiv180204082M}. Isolated real-time communications buses difficult the communication with with sensors and actuators inside the robot. In predictive maintenance the components need to be monitorized in real-time to detect possible faults or simply to know the condition of the components. A direct communication with the monitoring systems would enhance the integration of the robot components in monitoring systems.\\  

In this work we aim to explore and characterize the use of Ethernet and particularly, TSN, in robotics. Section \ref{related_work} will introduce an overview of the industrial communications available in the context of robotics and the corresponding related work. Section \ref{TSN} will present an analysis of TSN standards and the Ethernet timing model. Section \ref{results} will discuss the experimental results obtained while evaluating the presented hypotheses. Section \ref{conclusions} presents our conclusions and future work.

\section{Related work}
\label{related_work}

\subsection{A historical overview of industrial communications}
\label{fieldbus}

%

In order to understand the great diversity of communications protocols used in robotics it is necessary to explain the history of industrial communication protocols. In fact, many robots in the manufacturing industry nowadays are based on Programmable Logic Controller (PLC) technology or they interface with PLCs.\\

\subsubsection{Serial-based field-busses}

 Historically, field-busses substituted direct point to point digital and analog connections. They reduced the number of wires and provided a better communication platform to communicate PLCs with sensors, actuators and other low-level devices. The first industrial communications were serial-based fieldbus protocols such as DeviceNet\cite{devicenet}, Modbus\cite{modbus}, PROFIBUS\cite{profibus} and CC-Link\cite{cclink}. These protocols are still very popular today, because they are simple, robust and fast enough for real-time communications. They typically use protocols such as Controller Area Network (CAN) or RS-485 as their physical and data link layers. However, as industry applications became increasingly complex, there was an growing demand for higher speed, higher bandwidth, connection distances and a higher number of connection nodes. Serial-based protocols could not meet these requirements and the industry saw Ethernet as a good candidate to substitute serial-based field-busses for certain applications.\\



\subsubsection{Ethernet-based field-busses gain popularity}

\begin{figure}[h!]
\centering
 \includegraphics[width=0.48\textwidth]{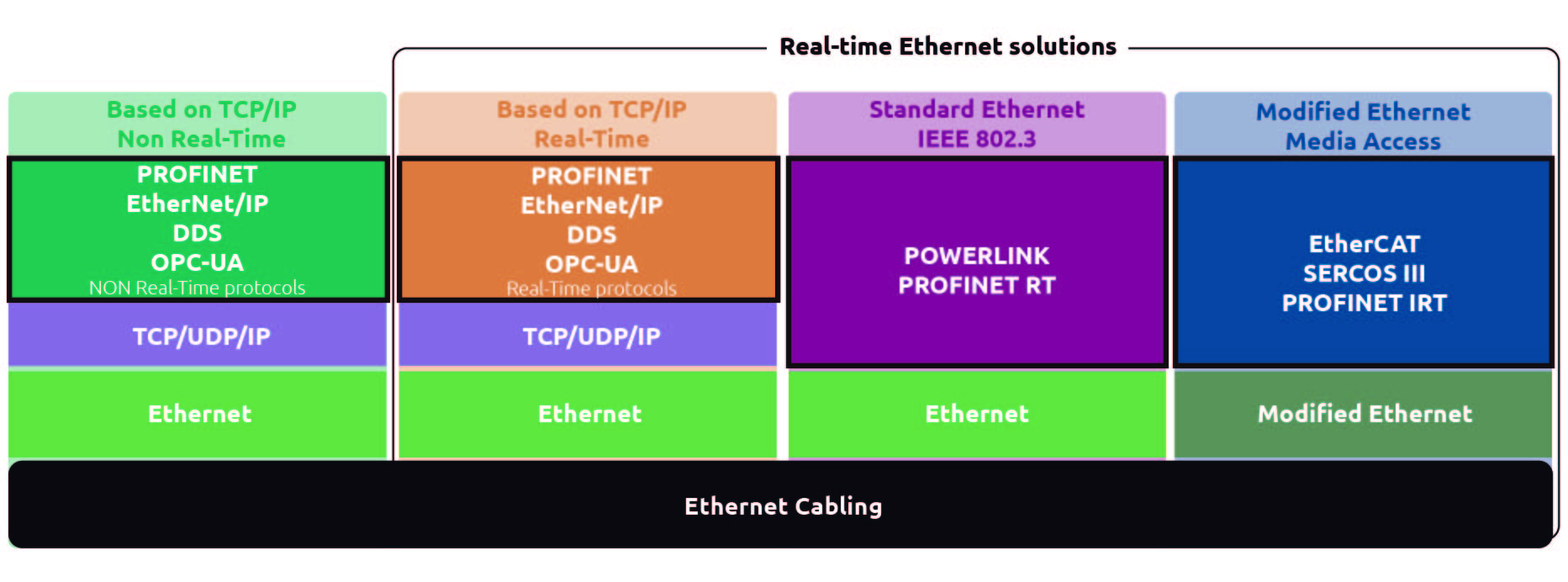}
\caption{\footnotesize Classification of real-time Ethernet solutions according to their network layers.}
\label{stack}
\end{figure}

In the 90's Ethernet was widely used in industry, but just for high level applications. It was not possible to use the commercial off-the-shelf Ethernet for time-critical applications such as distributed motor control. The problem was the intrinsic lack of determinism of the TCP/IP layer and the non-deterministic CSMA/CD algorithm to deal with packet collisions. The consequence of this was that different manufacturers developed different solutions for the same problem. For applications with low real-time requirements, high level protocols were developed on top of the standard TCP/IP and UDP/IP layers (Ethernet/IP\cite{ethernetIP} or PROFINET\cite{profinet}). For applications with higher real-time requirements, the TCP/IP or UDP/IP layers were substituted by a custom stack, to achieve higher determinism (POWERLINK\cite{powerlink}, PROFINET RT). For high critical applications where hard-real time was required, for example field devices such as sensors and actuators, some manufacturers developed vendor specific protocols based on a modified Ethernet technology (EtherCAT\cite{etherCAT}, SERCOS III\cite{sercos}, PROFINET IRT). A survey by Felser\cite{IEC_standards} in 2005 presented a further classification of real-time Ethernet solutions based on the layer where modifications are introduced. This classification is illustrated in Figure \ref{stack}.

Each protocol has been specialized in certain tasks and their use depends highly on the application. Because of this specialization, one protocol is not capable to meet all the requirements of a complex application, and at the end, more than one protocol are usually combined. Communications scene is a complex collection of technologies which lead to a high number of interoperability problems and an increase of the integration efforts. Despite of some standardization efforts by the the International Electrotechnical Commission (IEC), these problems were not solved and remain today. As stated in \cite{IEC_standards}, the International Electrotechnical Commission (IEC) failed to define one standard Real Time Ethernet (RTE) solution. Consequently, the set of standards finally defined more than a dozen of technical different solutions.\\ 




In robotics, the lack of a real standard protocol burdens the component integration or robot to robot communications. For example, end-effector manufacturers need to integrate different communication protocols in their products, depending on the protocols that a robot manufacturer supports. The list of industrial protocols used by the robot manufacturers is considerable. In many cases, and depending on the manufacturer, some industrial protocols are supported while some others not (see table \ref{protocols_table}).\\

\begin{table*}[]
\centering
\caption{\footnotesize Communication protocols used by some of the main Industrial robot manufacturers.}
\label{protocols_table}
\begin{tabular}{cccccccccc}
\cline{2-9}
\multicolumn{1}{c|}{}                           & \multicolumn{8}{c|}{Industrial protocols}                                                                                                                                                                                                                               &                              \\ \cline{2-9}
\multicolumn{1}{l|}{}                           & \multicolumn{4}{c|}{\cellcolor[HTML]{FFFFFF}\textbf{Serial based protocols}}                                                       & \multicolumn{4}{c|}{\cellcolor[HTML]{FFFFFF}\textbf{Ethernet based protocols}}                                                     & \multicolumn{1}{l}{}         \\ \hline
\multicolumn{1}{|c|}{Manufacturer}              & \multicolumn{1}{c|}{DeviceNET} & \multicolumn{1}{c|}{CC-Link} & \multicolumn{1}{c|}{Profibus-DP} & \multicolumn{1}{c|}{Modbus RTU} & \multicolumn{1}{c|}{MODbus TCP} & \multicolumn{1}{c|}{Ethernet/IP} & \multicolumn{1}{c|}{Profinet} & \multicolumn{1}{c|}{EtherCAT} & \multicolumn{1}{c|}{I.R.*}    \\ \hline
\multicolumn{1}{|c|}{\textbf{FANUC}}            & \multicolumn{1}{c|}{Yes}       & \multicolumn{1}{c|}{Yes}     & \multicolumn{1}{c|}{Yes}         & \multicolumn{1}{c|}{No}         & \multicolumn{1}{c|}{Yes}        & \multicolumn{1}{c|}{Yes}         & \multicolumn{1}{c|}{Yes}      & \multicolumn{1}{c|}{No}       & \multicolumn{1}{c|}{400.000} \\ \hline
\multicolumn{1}{|c|}{\textbf{Yaskawa}}          & \multicolumn{1}{c|}{Yes}       & \multicolumn{1}{c|}{Yes}     & \multicolumn{1}{c|}{Yes}         & \multicolumn{1}{c|}{Yes}        & \multicolumn{1}{c|}{No}         & \multicolumn{1}{c|}{Yes}         & \multicolumn{1}{c|}{No}       & \multicolumn{1}{c|}{No}       & \multicolumn{1}{c|}{360.000} \\ \hline
\multicolumn{1}{|c|}{\textbf{ABB}}              & \multicolumn{1}{c|}{Yes}       & \multicolumn{1}{c|}{Yes}     & \multicolumn{1}{c|}{Yes}         & \multicolumn{1}{c|}{No}         & \multicolumn{1}{c|}{No}         & \multicolumn{1}{c|}{Yes}         & \multicolumn{1}{c|}{Yes}      & \multicolumn{1}{c|}{No}       & \multicolumn{1}{c|}{300.000} \\ \hline
\multicolumn{1}{|c|}{\textbf{KAWASAKI}}         & \multicolumn{1}{c|}{Yes}       & \multicolumn{1}{c|}{Yes}     & \multicolumn{1}{c|}{Yes}         & \multicolumn{1}{c|}{No}         & \multicolumn{1}{c|}{Yes}        & \multicolumn{1}{c|}{Yes}         & \multicolumn{1}{c|}{No}       & \multicolumn{1}{c|}{No}       & \multicolumn{1}{c|}{110.000} \\ \hline
\multicolumn{1}{|c|}{\textbf{Denso}}            & \multicolumn{1}{c|}{Yes}       & \multicolumn{1}{c|}{Yes}     & \multicolumn{1}{c|}{Yes}         & \multicolumn{1}{c|}{No}         & \multicolumn{1}{c|}{No}         & \multicolumn{1}{c|}{Yes}         & \multicolumn{1}{c|}{Yes}      & \multicolumn{1}{c|}{No}       & \multicolumn{1}{c|}{95.000}  \\ \hline
\multicolumn{1}{|c|}{\textbf{KUKA}}             & \multicolumn{1}{c|}{Yes}       & \multicolumn{1}{c|}{No}      & \multicolumn{1}{c|}{Yes}         & \multicolumn{1}{c|}{No}         & \multicolumn{1}{c|}{No}         & \multicolumn{1}{c|}{Yes}         & \multicolumn{1}{c|}{Yes}      & \multicolumn{1}{c|}{Yes}      & \multicolumn{1}{c|}{80.000}  \\ \hline
\multicolumn{1}{|c|}{\textbf{Stäubli}}          & \multicolumn{1}{c|}{Yes}       & \multicolumn{1}{c|}{Yes}     & \multicolumn{1}{c|}{Yes}         & \multicolumn{1}{c|}{No}         & \multicolumn{1}{c|}{Yes}        & \multicolumn{1}{c|}{Yes}         & \multicolumn{1}{c|}{Yes}      & \multicolumn{1}{c|}{Yes}      & \multicolumn{1}{c|}{45.000}  \\ \hline
\multicolumn{1}{|c|}{\textbf{Universal Robots}} & \multicolumn{1}{c|}{No}        & \multicolumn{1}{c|}{No}      & \multicolumn{1}{c|}{No}          & \multicolumn{1}{c|}{No}         & \multicolumn{1}{c|}{Yes}        & \multicolumn{1}{c|}{No}          & \multicolumn{1}{c|}{No}       & \multicolumn{1}{c|}{No}       & \multicolumn{1}{c|}{20.000}  \\ \hline
\multicolumn{10}{l}{*I.R. Number of installed robots}                                                                                                                                                                                                                                                                                                    
\end{tabular}
\end{table*}

\subsubsection{Bringing real-time determinism on standard Ethernet}


As a result of an increasing need of real-time capabilities in standard Ethernet, the TSN task group was created. Previously named as Audio/Video Bridging Task Group, its main goal was to achieve time-synchronized low latency streaming services through IEEE 802 networks. Later, because of the interest of industry, the goal was extended to define mechanisms for the time-sensitive transmission of data over Ethernet networks. One of the main goals of TSN was to provide a unified layer 2 for real-time communications, so different vendors are compatible among them and allowing at the same time the convergence of traffic and real-time communications.\\

An important advantage of communications based on standard Ethernet is that these protocols will be automatically benefited by all the features and improvements that the standard provides. For example, it is possible to use higher link capacities such as 10 Gbps or higher as the technology improves. It also becomes easier to integrate communication frameworks and common tools, simply because most of the software has been developed on top of TCP and UDP protocols. In our experience, it can be challenging and time consuming to integrate robotics frameworks with field-busses such as CAN or EtherCAT. Another important advantage of standard Ethernet based communications is related to costs, because of the high number of vendors that offer Ethernet based devices, the technology cost is lower compared to vendor specific solutions.\\

\subsubsection{Network protocols for robotics}

We claim that TSN-based protocols are going to replace existing legacy protocols. However, in the short term, existing protocols will coexist and slowly converge towards TSN, which will provide better interoperability characteristics. Some of the existing industrial protocols based on Ethernet will be benefited by TSN. For example, Profinet and Ethernet/IP. But also, TSN will open the doors for other protocols and frameworks which were not originally destined for hard real-time communications, such as OPC Unified Architecture (OPC-UA)\cite{opcua} and Data Distribute Service (DDS)\cite{dds}. The Internet Industrial Consortium (IIC) selected these frameworks as two out of four connectivity frameworks standards. As stated in `The Industrial Internet of Things Volume G5: Connectivity Framework' \cite{IICframework}, field-busses implement parts of the connectivity transport and framework functions, but none of them do satisfy all of the connectivity core standard criteria. Both OPC UA and DDS vendors are currently working to improve the performance and real-time capabilities by integrating TSN in their frameworks. \\

OPC UA and DDS are increasingly used in robotics. Their appliance depends highly on the industrial area. OPC UA has been selected as the backbone of Industry 4.0 and seems to be gaining traction in the industrial manufacturing robotics market. On the other hand, DDS is widely used in military and professional robotic industries (healthcare, warehouse automation, aerospace, etc.) More recently, DDS is also being applied in the automotive and consumer robotics markets. Besides the traditional manufacturing industry, our team observes that DDS is gaining presence in the overall robotics landscape. Related to this, DDS has been selected as the official communication middleware for the Robot Operating System 2 (ROS 2)\cite{ros2, ros2design} framework, the de facto standard for robot application development. In the coming years, we can expect that both OPC UA and DDS will extend their usage in robotics helping roboticists achieve more hard real-time robotic applications thanks to TSN. In addition, we foresee a wider integration of these frameworks (or extensions of them) in resource constrained robotic components.



\section{Time Sensitive Networking}
\label{TSN}
\subsection{Overview of the standards}


TSN is composed by a set of standards that aim to make Ethernet more deterministic. Several of these standards treat the problem of how to achieve bounded latencies. As we explained in a previous section (\ref{fieldbus}), standard Ethernet did not meet the deterministic capabilities for real-time applications. In the past, Ethernet endpoints used to be half-duplex and used hubs for the connection. One of the first problems Ethernet faced was the collision domain problem. Data packets could collide with one another while being sent. To deal with this, an arbitration algorithm called Carrier Sense Multiple Access with Collision Detection (CSMA/CD) was used. With the introduction of full duplex Ethernet switches, the network communications were isolated in different domains, so they do not contend for the same wire. One of the major issues was solved, but the main source causing a lack of determinism got moved to congestion problems inside the switch queues.\\ 


The main source of indeterminacy in nowadays switches is due to traffic contention to access the media access control (MAC) level. To limit and control the congestion problems at the queues, the IEEE introduced quality of service (QoS) mechanisms, such as traffic prioritization. Packet prioritization was introduced by the IEEE 802.1P task group in the 90's. This QoS technique, also known as class of service (CoS) consist in a 3-bit field called the Priority Code Point (PCP) within an Ethernet frame header when using VLAN tagged frames. This field allows to specify a priority value between 0 and 7 that can be used to prioritize traffic at the MAC layer.\\

The Ethernet switch may have one or more transmission queues for each bridge port. Each queue provides storage for frames that await to be transmitted. The frames will be assigned to each queue according their CoS. The switching transmission algorithm will then select the next packet of the queue with the highest priority. Frames stored with lower priority will be transmitted only if higher order queues are empty during the selection process.\\

\begin{figure}[h!]
\centering
 \includegraphics[width=0.3\textwidth]{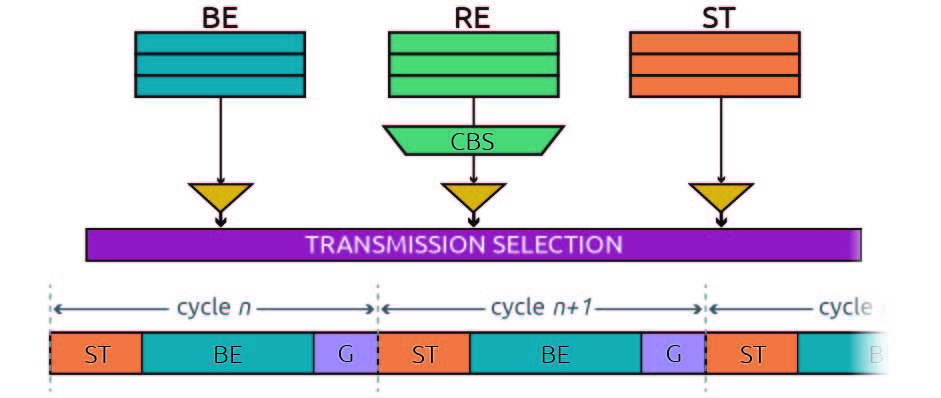}
\caption{\footnotesize Tranmission selection of the TSN Time-Aware Shaper.}
\label{TAS}
\end{figure}

Despite packet prioritization being an effective technique to decrease the traffic interference, it is still not enough to guarantee a deterministic latency. One of the problems is that lower priority traffic still interferes the higher priority traffic. That is, high priority frames will need to wait until lower priority ones finish their transmission. The delay will depend on the size of the packet being transmitted and on the number of switches along the frame path. In applications with low latency requirements and a high number of hops, the problem becomes very significant. Additionally, as Ethernet is asynchronous, the high priority frames sharing the same link can contend between them. To solve this problem, the 802.1 TSN task group developed time aware traffic scheduling, defined in 802.1Qbv \cite{7440741}. The Time-Aware Scheduler (TAS) is based on a TDMA (Time Division Multiple Access) to divide a cycle time into time slots dedicated to a specific CoS. The TAS uses transmission gates for each queue and the gate can open or close the transmission of that queue (figure \ref{TAS}). The transmission selection algorithm selects the next frame of the higher order queue, but just from those queues with the gates opened. To prevent the MAC being busy when the scheduled frames arrive, the TAS introduces a guard band (G) in front of every time sensitive traffic time slice. This ensures access without delay to the MAC for time critical traffic. 

The gates are programmed specifying a cycle time and a gate control list. The list configures the times slices to open and close the gates of each queue. The gate control mechanism requires time synchronization among all the Time-Aware devices on the TSN network. The most used time synchronization protocol is the IEEE 1588 Precision Time Protocol (PTP), which synchronizes the clocks by exchanging Ethernet frames with synchronization information. The IEEE 802.1 TSN task group is working on a revision of IEEE 802.1AS \cite{5741898}, a profile of IEEE 1588 for audio/video systems. The new revision will add some characteristics needed in other fields, such as industrial control. One of the main requirements of IEEE 802.1AS capable devices in the TSN network is a sub-microsecond synchronization among them.\\

Another standard developed for deterministic communications is the IEEE 802.1Qbu \cite{7553415} which provides a frame preemption mechanism. This standard allows a high priority frame to interrupt a low priority frame in transmission. In order to decide if the frame can be preempted, the preemption mechanism needs a minimum necessary fragment of the frame. This fragment can be, in the worst case, 124 bytes. This means that using preemption solely does not guarantee an end-to-end deterministic latency. According to the standard, 802.1Qbu can be used in isolation to reduce latency and jitter or in combination with scheduling. When used with scheduling, it minimizes the protected window or guard band so the available bandwidth for preemptable traffic is optimized. Bandwidth optimization seems to be the main purpose of 802.1Qbu, which can be relevant for 100 Mbps.

\subsection{Switched Ethernet timing model}
\label{model}

In this section we analyze the delays involved in a switched Ethernet network. We will define an analytical model to determine the end-to-end latency in a linear topology compounded by bridged-endpoints in cut-trough mode. This model will help us to understand better the non-determinism sources of the end-to-end latency, and we will use it to analyze the results of the experiments in section \ref{results}.\\

The proposed model for the end-to-end latency is based on the delays defined in \cite{Kurose:2012:CNT:2584507} and \cite{shaper_vehicle}. First, we define some of the terms used to generalize an equation for the timing model: \\


\begin{itemize}
    \item \textbf{Frame transmission delay ($d_{t}$)}: time required to transmit all of the packet’s bits into the link.
    \item \textbf{Propagation delay ($d_{l}$)}: time for one bit to propagate from source to destination at propagation speed of the link.
    \item \textbf{Switch delay ($d_S$)}: time for one bit to traverse from the switch input port to the switch output port. 
    \item \textbf{Switch input delay ($d_{S_{in}}$)}:  delay of the switch ingress port, including the reception PHY and MAC latency.
    \item \textbf{Switch output delay ($d_{S_{out}}$)}:  delay of the switch egress port, including the transmission PHY and MAC latency.
    \item \textbf{Switch processing delay ($d_{S_{p}}$)}: time required to examine the packet’s header and determine where to direct the packet is part of the processing delay.
    \item \textbf{Switch queuing delay ($d_{S_{q}}$)}: time until a frame waits in the egress port of a switch to start the transmission onto the link.
   
\end{itemize}


\begin{figure}[h!]
\centering
 \includegraphics[width=0.3
\textwidth]{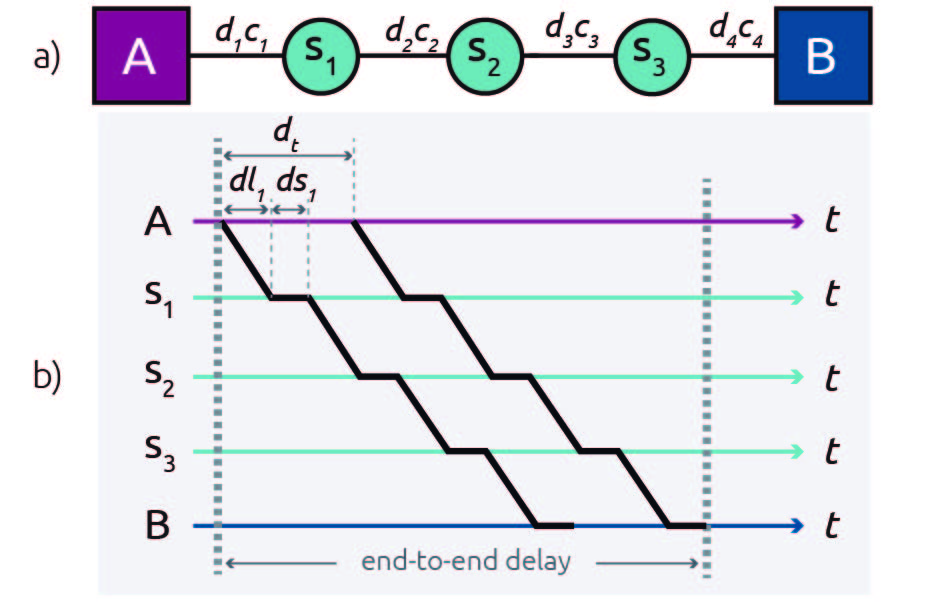}
\caption{\footnotesize a) Network topology for the timing model. b) Space-time diagram for end-to-end delay.}
\label{diagram}
\end{figure}

The propagation delay depends on the distance $d$ between two switches and the propagation speed of the link $s$. The transmission delay depends on the packet length $L$ and the link capacity $C$.

\begin{equation}
d_{l} = \frac{d}{s}
   \quad\mathrm{, }\quad 
d_{t} = \frac{L}{C}
\end{equation}

In cut-trough the switch delay does not depend on the packet length and can be expressed as shows equation \ref{eq:1}.

\begin{equation} \label{eq:1}
d_S(t) = d_{S_{in}} + d_{S_{p}} + d_{S_{out}} + d_{S_{q}}(t)
\end{equation}


The end-to-end delay from the source endpoint \textbf{A} to the destination endpoint \textbf{B} can be expressed as the sum of the delays of all the switches and links in the path, being $n$ the number of links and $n-1$ the number of switches along the path.

\begin{equation} \label{eq:2}
d_{AB}(t) = d_{t_1} + \sum_{i=1}^n \Big( d_{l_i} \Big) + \sum_{i=1}^{n-1} \Big( d_{S_i}(t)  \Big)  
\end{equation}

The key idea is that all the terms of equation \ref{eq:2} are deterministic except the queuing delay, which depends on the switch queue occupancy when the frame arrives at time $t$. This delay is the main problem to bound the latency in switched Ethernet. There are different ways to bound this delay, for example QoS techniques such as Weighted Fair Queueing (QFQ) or Strict priority \cite{WFQ}, but there is still certain delay and jitter which limits the real-time performance. Using a TSN time aware shaper and a appropriate schedule, this delay can be completely eliminated. Once this delay disappears, the end-to-end latency becomes deterministic and, combined with cut-trough, it is possible to achieve a very low latency.\\

In the context of robotics, for a simple robot manipulator, the worst case end-to-end latency from the actuators to the robot control will determine the minimum achievable control cycle time and the maximum number of actuators allowed for a fixed cycle time. This is why reducing the queuing delay is critical to achieve low cycle times.\\

The cycle time is an important metric to measure the performance of the communication protocol. It can be defined as the time necessary to exchange the input and output data between the controller and all the sensors and actuators. The TAS provides real-time performance, which makes Ethernet with TSN comparable in terms of real-time performance with other real-time communication protocols. Jasperneite et al.\cite{4416748} introduced an analytic method to estimate input and output cycle times for Ethernet technologies. They used this method to compare EtherCAT with Profinet IRT.\\

Bernier\cite{robotiqBernier} extended the method to Modbus/TCP solution and Ethernet/IP. In his work, he defines the cycle time as the time necessary to exchange the input and output data between the controller and all the sensors and actuators. Bruckner et al.\cite{opcUATSN} used this method to compare the most common communication technologies performance with the OPC UA over TSN. The conclusions drawn by this last work shows how a TSN based technology using 1 Gbit Ethernet and a frame aggregation approach can outperform other hard real-time industrial protocols. In the following section, we will challenge these results experimentally in the context of robotics.

\section{Experimental results}
\label{results}

\begin{figure}[h!]
\centering
 \includegraphics[width=0.45\textwidth]{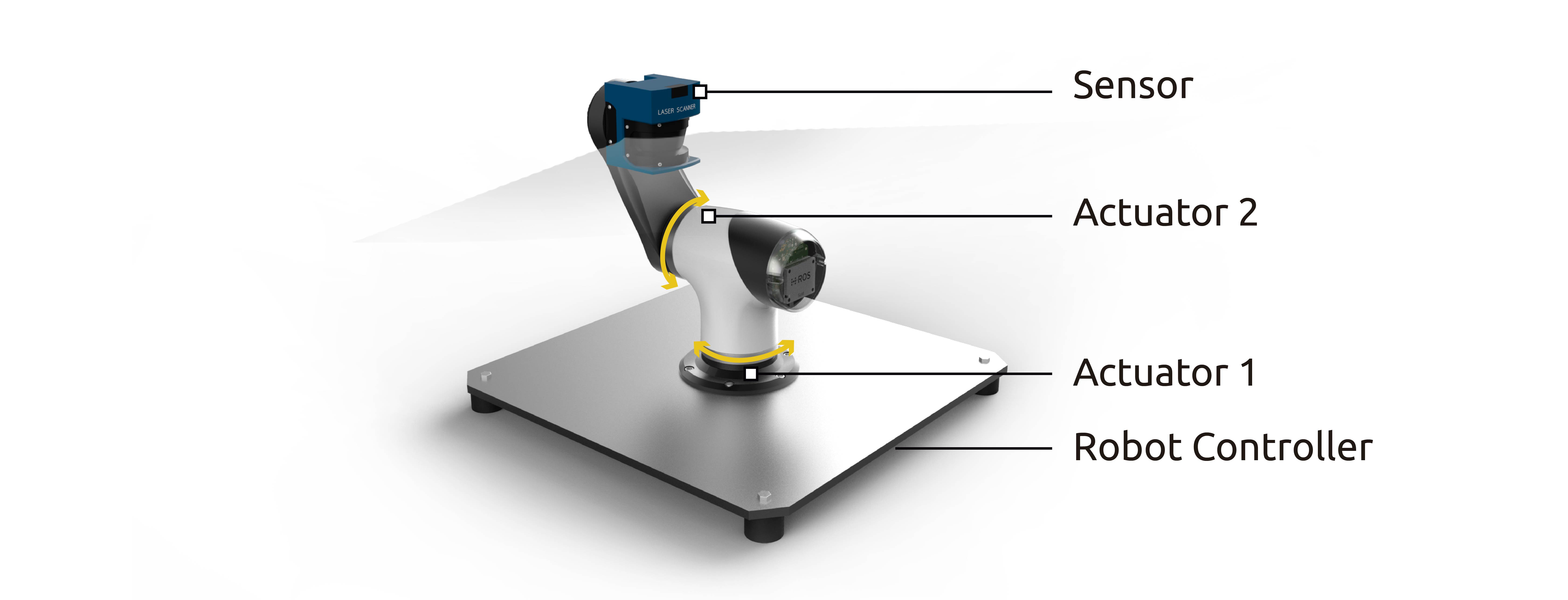}
\caption{\footnotesize Exemplary modular robotic arm used for the experimental setup.}
\label{robot_arm}
\end{figure}

For the experimental setup, we have selected a typical robotic use case with mixed-critical traffic. In particular, we have chosen a modular robotic arm with a high bandwidth sensor attached at the end (Figure \ref{robot_arm}). Such robot arm is an interesting use case for TSN because it gathers traffic with different characteristics. The actuators require hard real-time low latency traffic while sensors, such as high resolution camera or a laser scanner, generate high volume data. The proposed setup contains two actuators, \textbf{A1} and \textbf{A2}, a sensor \textbf{S} and a robot controller \textbf{RC}.\\ 
 



 
\begin{figure}[h!]
\centering
 \includegraphics[width=0.45\textwidth]{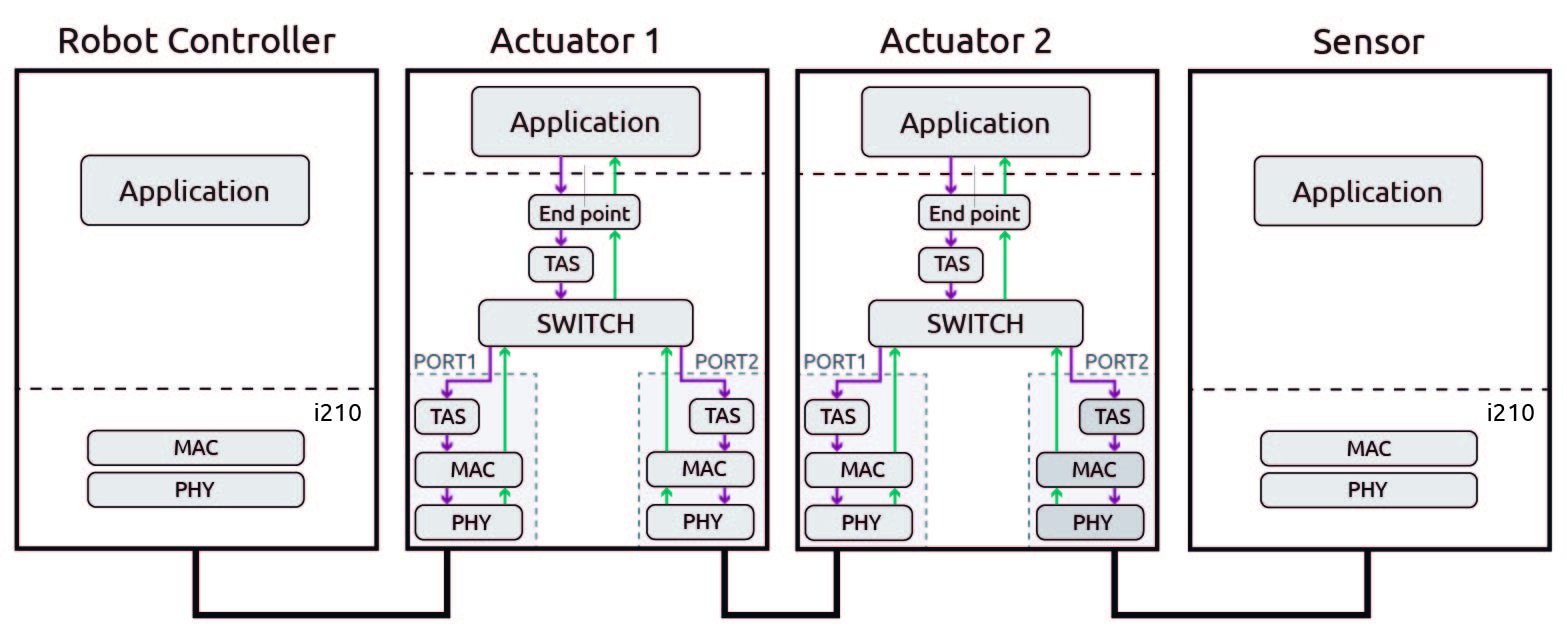}
\caption{\footnotesize Experimental setup networking devices overview.}
\label{setup}
\end{figure}

 
For the experiment we will use two TSN capable bridged endpoints and two PCs, each one with an Intel i210 card (figure \ref{setup}). The TSN bridged endpoints simulate the actuators and the PCs the sensor \textbf{S} and the robot controller \textbf{RC}. The sensor \textbf{S} is connected to the last actuator and will send a high bandwidth to the robot controller. As the sensor is connected in a chain topology, the sensor traffic goes through each of the actuators switches. This traffic will contend with the time sensitive traffic from the actuators switch queues, which will produce a queuing delay.



\begin{figure}[h!]
\centering
 \includegraphics[width=0.45\textwidth]{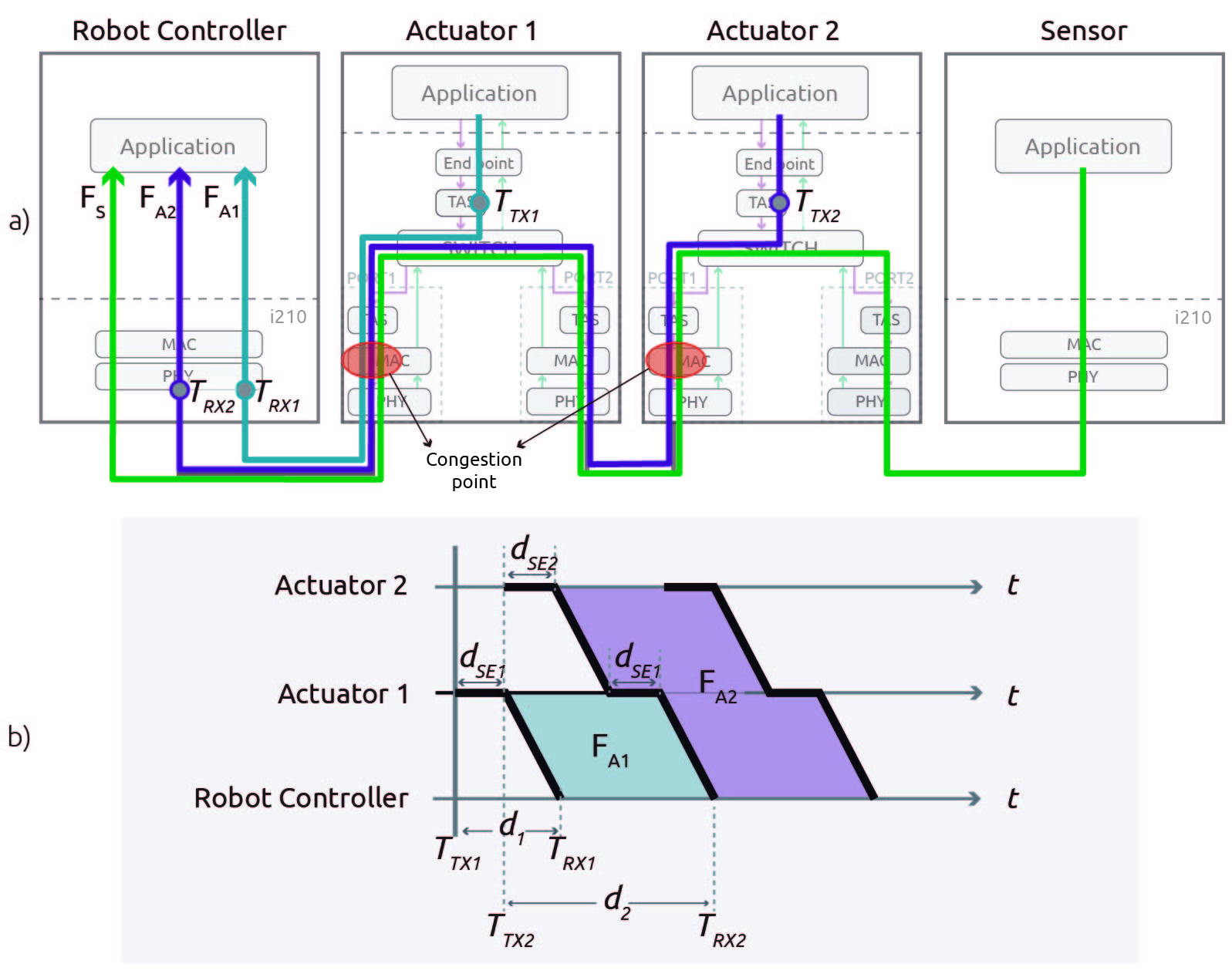}
\caption{\footnotesize a)Experimental setup traffic flow. b)End-to-end delay space-time diagram.}
\label{setup_flow}
\end{figure}

We call $F_{A1}$ to the traffic flow from \textbf{A1} to \textbf{RC}, $F_{A2}$ to the traffic flow from \textbf{A2} to \textbf{RC} and $F_{S}$ to the traffic flow from \textbf{S} to \textbf{RC}, as shown in figure \ref{setup_flow}.  The measurements are performed in \textbf{RC} using the hardware time-stamps capabilities of the i210 network card. For the transmission time, we take advantage of the TSN capabilities of the bridged endpoints. We set the transmission times $t_{TX_1}$ and $t_{TX_2}$ of the TSN bridged endpoints by configuring the TAS of each endpoint. For an actuator $i=1,2$, we calculate the delay $d_{i}$ from \textbf{$A_i$} to \textbf{RC} by a simple subtraction.



\begin{equation} \label{eq:3}
d_{i} = t_{RX_i} - t_{TX_i}
\end{equation}

As shown in figure \ref{setup_flow}, the hardware time-stamp of \textbf{RC} measures the arrival time of the Start of Frame Delimiter (SFD). This means that the frame transmission delay is not included in our measurements. As the switches are configured in cut-trough, the transmission delay is neither included in the switch delay and hence the measured delay does not depend on the frame length. From \ref{setup_flow} we can relate the measurements with the delays expressed in equation \ref{eq:2}. Notice that in this particular case the sources are switched endpoints. The delays from the endpoint trough the switch $d_{SE_i}$ are not the same of the switch delay $d_{S_i}$  because it does not include the input port delay. Expressing the measured delay in terms of the switch delay, we arrive to the expressions \ref{eq:4} and \ref{eq:5} for $d_1$ and $d_2$ respectively.

\begin{equation} \label{eq:4}
\begin{split}
d_{1}(t) & = d_{SE_1}(t) +d_{l_1} \\
              & = d_{S_{p1}} + d_{S_{out1}} + d_{S_{q1}}(t) + d_{l_{1}}\\
              & = K_1 + d_{S_{q1}(t)}
\end{split}              
\end{equation}

\begin{equation} \label{eq:5}
\begin{split}
d_{2}(t) & = d_{SE_2}(t) +d_{l_2} + d_{S_1}(t) +d_{l_1} \\
         & = ...\\
         & = K_2 + d_{S_{q1}}(t) + d_{S_{q2}}(t)
\end{split}              
\end{equation}

To generate the $F_{A1}$ and $F_{A2}$ traffic, we have used UDP/IP with 256 Bytes payload at a 1 ms rate. For $F_{S}$, we used the \textit{iperf} tool to generate a network load using a 1500 Byte payload with a traffic bandwidth of 900 Mbps for 1 Gbps link capacity and 90 Mbps for 100 Mbps link capacity. For the calculations, we measured 10000 samples during over a 10 seconds period.\\


\subsubsection*{Experiment 1. Same priority blocking}

In this experiment, we measure the blocking effect of same priority traffic. All the traffic flows are sent trough lowest priority queue, which is the best-effort queue (BE). $F_{A1}$ and $F_{A2}$ will contend with $F_{S}$ in the port 1 BE queue of the $A1$ and $A2$ switches. The blocking delay depends on the implemented transmission selection mechanism of the switch. The switches used for this setup use a FIFO occupancy credit based arbitration, from now on referred as CBF. CBF prioritizes the traffic coming from a higher occupancy ingress port queue. This makes the blocking effect highly dependent on the sending rates of the contending sources of traffic.\\  

The results (Figure \ref{a1_be}) show worst case delays in the order of milliseconds, which clearly shows that the latency is unbounded. Increasing $F_{S}$ bandwidth or decreasing $F_{A1}$ sending rates, we get higher delays and even packet loss. Note that the purpose of this experiment is not to analyze the queuing delay of the CBF but only to illustrate the non-determinism of this arbitration policy due to the same priority blocking. \\

\subsubsection*{Experiment 2. Lower priority blocking}

In this experiment, we measure the blocking effect of lower priority traffic using a strict priority transmission. $F_{A1}$ and $F_{A2}$ are sent trough the highest priority queue of the switches, which is the scheduled traffic queue (ST). Previous work \cite{802.1p} analyzed the effect of lower priority blocking. The worst case delay added by each switch is given by the frame transmission time for the maximum frame size allowed by the switch (MTU). For a 1500 Byte MTU and a 100 Mbps link capacity, the worst case added delay would be 120 $ \mu s$, for 1 Gbps 12 $ \mu s$. \\

\begin{figure*}[h!]
  \begin{subfigure}[t]{.5\textwidth}
    \centering
    \includegraphics[width=0.8\linewidth]{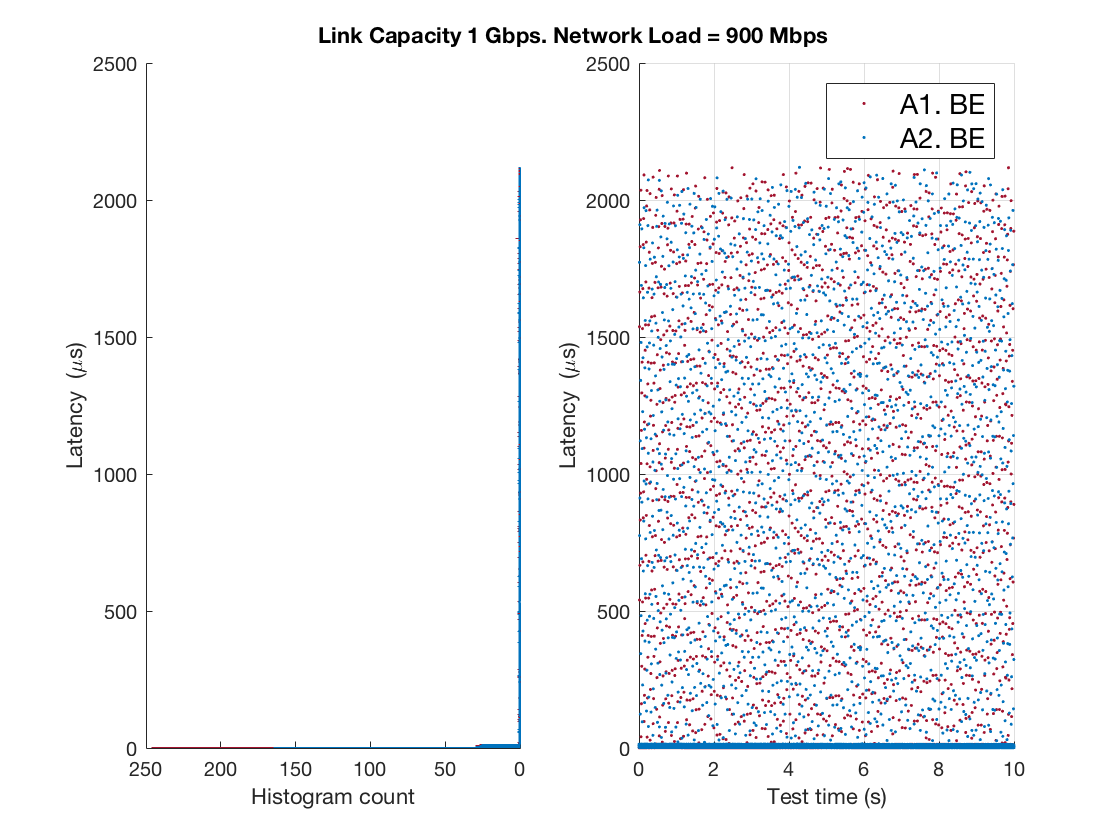}
    \caption{ }
    \label{a1_be}
  \end{subfigure}
  \hfill
  \begin{subfigure}[t]{.5\textwidth}
    \centering
    \includegraphics[width=0.8\linewidth]{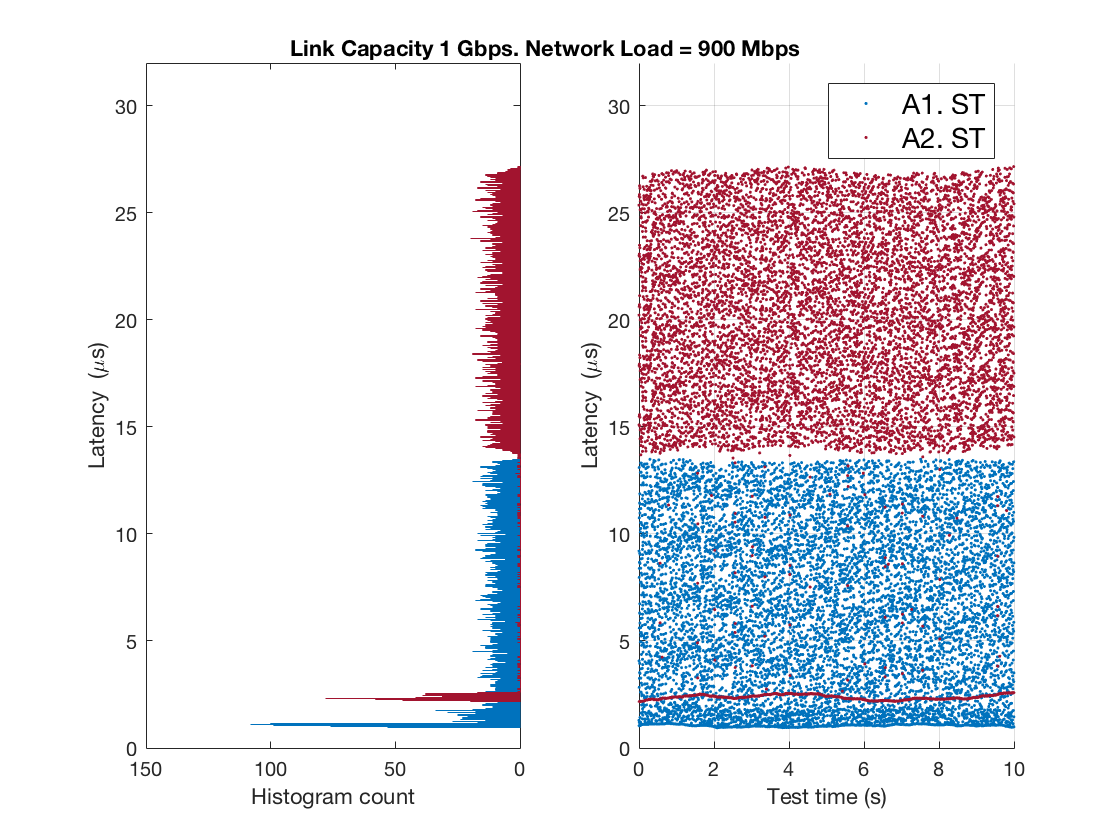}
    \caption{}
    \label{a1_st}
  \end{subfigure}

  \medskip

  \begin{subfigure}[t]{.5\textwidth}
    \centering
    \includegraphics[width=0.8\linewidth]{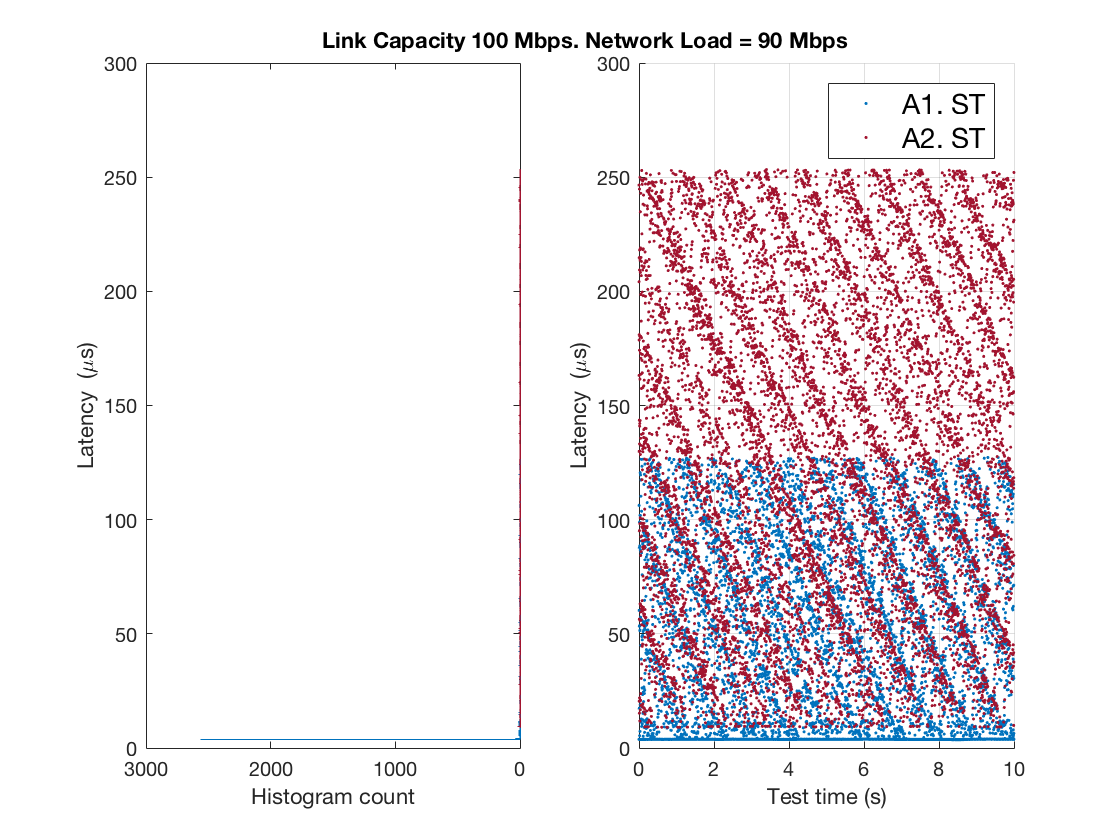}
    \caption{}
    \label{a2_st}
  \end{subfigure}
  \hfill
  \begin{subfigure}[t]{.5\textwidth}
    \centering
    \includegraphics[width=0.8\linewidth]{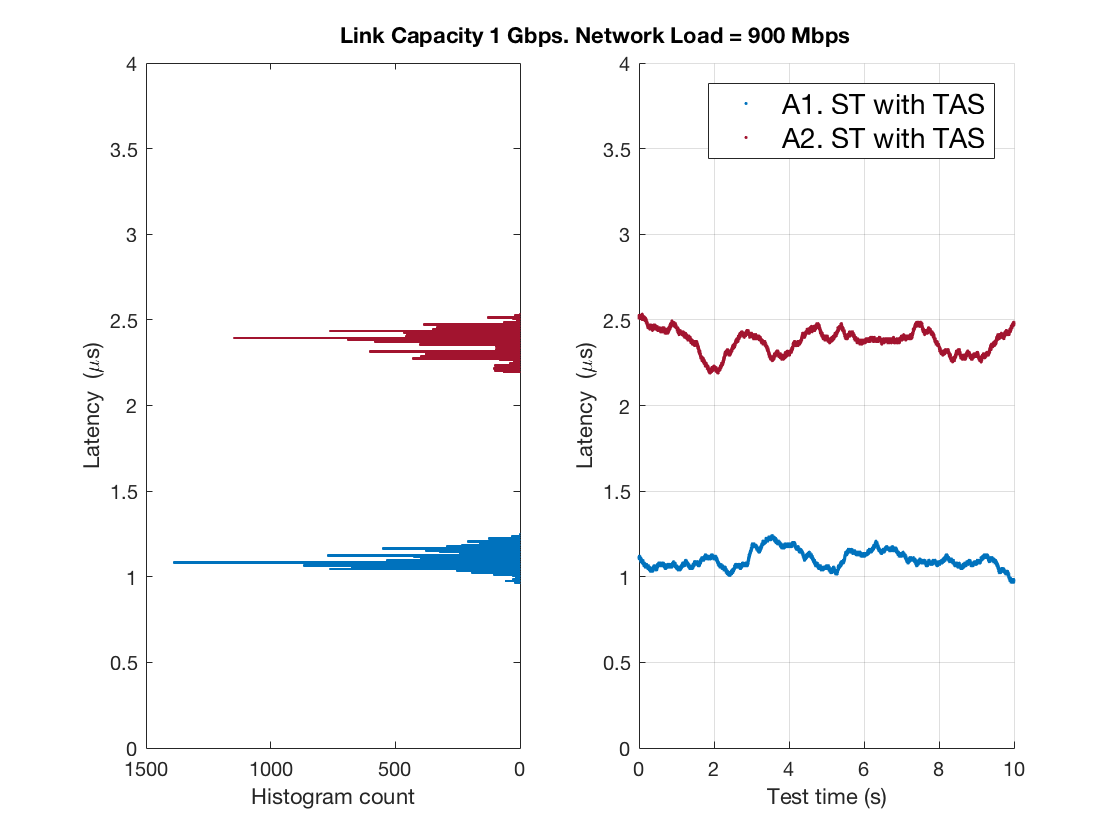}
    \caption{}
    \label{a1_a2_tas}
  \end{subfigure}
  \caption{ \footnotesize Timeplot delay measurements for 10s.Terms: C=Link Capacity,. Terms: \textcolor[rgb]{0,0,1}{A1=Actuator1 (Blue)}, \textcolor[rgb]{1,0,0}{A2=Actuator 2 (Red)}, BE=Best-effort queue, ST=Scheduled Traffic queue. a) Same priority blocking for C=1 Gbps b)  Lower priority blocking for C=1 Gbps c) Lower priority blocking for C=100 Mbps  d) Using a TAS for C=1 Gbps}
\end{figure*}

The experimental results are presented in Tables \ref{100mbps_table} and \ref{1gpbs_table} and pictured in Figures \ref{a1_st} and \ref{a2_st}. These results confirm the expected worst case delay. In the case of 100 Mbps, we have a maximum delay of 127.22 $ \mu s$ for $d_1$ and 253.33 $ \mu s$ for $d_2$. The results show how the effect is accumulative for each bridge. 
While this queuing delay in this case has an upper bound, it clearly highly limits the real-time performance and the scalability of the system. For 100 Mbps and network load, the delay goes approximately from 3 to 120 $ \mu s$ traffic for $A_1$ and from 9 to 250 $ \mu s$ for $A_2$. As $F_{A1}$ and $F_{A2}$ are sent asynchronously, $F_{A1}$ and $F_{A2}$ can contend between them. Therefore, in this scenario, apart from the lower priority blocking, the worst case must take into account the same priority blocking of $F_{A1}$ and $F_{A2}$.\\ 


\subsubsection*{Experiment 3. Using a TSN Time-Aware Shaper}

In this experiment we configure the TAS of $A_1$ and $A_2$ switches egress ports. As all the TAS are synchronized with sub-microsecond accuracy, the scheduled traffic is perfectly isolated from lower priority traffic and from the same priority traffic. As shown in Figure \ref{a1_a2_tas}, the end-to-end latency is highly deterministic, the results with and without lower priority traffic are in the same order and sub-microsecond jitter is achieved.

\section{Conclusion}
\label{conclusions}

In this work we have presented an experimental setup to show the suitability of TSN for real-time robotic applications. We have compared the delays experienced by the queuing delay in Ethernet switches for standard Ethernet, against the delays when using a TSN Time-Aware shaper. The results showed the indeterminacy of Ethernet and how these problems can limit the scalability and performance in real-time robotic applications such as the exemplary modular robot. When the TSN Time-Aware shaper was used, the results showed that the time sensitive traffic was perfectly isolated from lower priority traffic, maintaining low latency and jitter even in the presence of high bandwidth background traffic. These results suggest that it is possible to develop hard real-time motion control systems mixed with high bandwidth sensors, such as lidars and high resolution cameras. \\

Based on the presented results, we claim that Ethernet with TSN standards will become the de facto standard for communications on layers 1 and 2, in robotics. We argue that, within robotics, many of the existing real-time industrial solutions will slowly be replaced by TSN. For higher layers, we foresee a contending landscape where the integration of TSN in different middleware solutions focused on interoperability such as OPC-UA and DDS promise to deliver a bottom-up real-time communication solution.






%

\bibliographystyle{IEEEtran}
\bibliography{references}


\onecolumn


\label{ap_figures}

\begin{table*}[]
\centering
\caption{\footnotesize Delay results for 100 Mpbs Link Capacity}
\label{100mbps_table}
\begin{tabular}{l|c|c|c|c|c|c|c|c|}
\cline{2-9}
                                          & \multicolumn{8}{c|}{Link Capacity 100 Mbps}                                                        \\ \cline{2-9} 
                                          & Queue               & QoS & Network load & Min($\mu$s) & Max($\mu$s)   & Mean($\mu$s) & Std($\mu$s)  & Max-Min($\mu$s) \\ \hline
\multicolumn{1}{|l|}{\multirow{6}{*}{A1}} & \multirow{2}{*}{BE} & \multirow{2}{*}{CBF} & -            & 3.61    & 10.04     & 3.74     & 0.12     & 6.8         \\ \cline{4-9} 
\multicolumn{1}{|l|}{}                    &                     &  & 90 Mbps      & 3.67    & 2,804.15  & 64.60    & 141.11   & 2,800.48    \\ \cline{2-9} 
\multicolumn{1}{|l|}{}                    & \multirow{4}{*}{ST} & \multirow{2}{*}{SP}  & -            & 3.74    & 3.98      & 3.83     & 0.045    & 0.24        \\ \cline{4-9} 
\multicolumn{1}{|l|}{}                    &                     &   & 90 Mbps      & 3.62    & 127.22    & 35.84    & 40.02    & 123.6       \\ \cline{3-9} 
\multicolumn{1}{|l|}{}                    &                     & \multirow{2}{*}{TAS} & -            & 3.58    & 4.06      & 3.82     & 0.0077   & 0.48        \\ \cline{4-9} 
\multicolumn{1}{|l|}{}                    &                     &  & 90 Mbps      & 3.66    & 3.98      & 3.82     & 0.076    & 0.32        \\ \hline
\multicolumn{1}{|l|}{\multirow{6}{*}{A2}} & \multirow{2}{*}{BE} & \multirow{2}{*}{CBF} & -            & 9.07    & 9.87      & 9.36     & 0.15     & 0.80        \\ \cline{4-9} 
\multicolumn{1}{|l|}{}                    &                     &  & 90 Mbps      & 9.15    & 22,822.12 & 189.00   & 1,381.40 & 22,812.96   \\ \cline{2-9} 
\multicolumn{1}{|l|}{}                    & \multirow{4}{*}{ST} & \multirow{2}{*}{SP}  & -            & 9.10    & 9.58      & 9.35     & 0.087    & 0.48        \\ \cline{4-9} 
\multicolumn{1}{|l|}{}                    &                     &   & 90 Mbps      & 9.01    & 253.33    & 133.99   & 69.78    & 244.33      \\ \cline{3-9} 
\multicolumn{1}{|l|}{}                    &                     & \multirow{2}{*}{TAS} & -            & 9.24    & 9.80      & 9.49     & 0.10     & 0.56        \\ \cline{4-9} 
\multicolumn{1}{|l|}{}                    &                     &  & 90 Mbps      & 9.16    & 9.64      & 9.43     & 0.08     & 0.48        \\ \hline
\end{tabular}
\end{table*}

\begin{table*}[]
\centering
\caption{\footnotesize Delay results for  1 Gbps Link Capacity}
\label{1gpbs_table}
\begin{tabular}{l|c|c|c|c|c|c|c|c|}
\cline{2-9}
                                          & \multicolumn{8}{c|}{Link Capacity 1 Gbps}                                                        \\ \cline{2-9} 
                                          & Queue               & QoS & Network load & Min($\mu$s) & Max($\mu$s)  & Mean($\mu$s) & Std($\mu$s) & Max-Min($\mu$s) \\ \hline
\multicolumn{1}{|l|}{\multirow{6}{*}{A1}} & \multirow{2}{*}{BE} & \multirow{2}{*}{CBF} & -            & 0.97    & 1.19     & 1.06     & 0.04    & 0.22        \\ \cline{4-9} 
\multicolumn{1}{|l|}{}                    &                     &  & 900 Mbps     & 0.91    & 2,114.14 & 213.23   & 496.22  & 2,114.14    \\ \cline{2-9} 
\multicolumn{1}{|l|}{}                    & \multirow{4}{*}{ST} & \multirow{2}{*}{SP}  & -            & 0.90    & 1.22     & 1.06     & 0.08    & 0.32        \\ \cline{4-9} 
\multicolumn{1}{|l|}{}                    &                     &   & 900 Mbps     & 0.94    & 13.48    & 6.35     & 3.93    & 12.54       \\ \cline{3-9} 
\multicolumn{1}{|l|}{}                    &                     & \multirow{2}{*}{TAS} & -            & 1.00    & 1.23     & 1.12     & 0.04    & 0.22        \\ \cline{4-9} 
\multicolumn{1}{|l|}{}                    &                     &  & 900 Mbps     & 0.96    & 1.24     & 1.10     & 0.05    & 0.27        \\ \hline
\multicolumn{1}{|l|}{\multirow{6}{*}{A2}} & \multirow{2}{*}{BE} & \multirow{2}{*}{CBF} & -            & 2.12    & 2.65     & 2.35     & 0.12    & 0.52        \\ \cline{4-9} 
\multicolumn{1}{|l|}{}                    &                     &  & 900 Mbps     & 2.14    & 2,118.02 & 216.23   & 492.59  & 2,115.88    \\ \cline{2-9} 
\multicolumn{1}{|l|}{}                    & \multirow{4}{*}{ST} & \multirow{2}{*}{SP}  & -            & 2.15    & 2.61     & 2.40     & 0.10    & 0.46        \\ \cline{4-9} 
\multicolumn{1}{|l|}{}                    &                     &   & 900 Mbps     & 2.16    & 27.16    & 18.55    & 6.49    & 24.99       \\ \cline{3-9} 
\multicolumn{1}{|l|}{}                    &                     & \multirow{2}{*}{TAS} & -            & 2.30    & 2.55     & 2.43     & 0.05    & 0.24        \\ \cline{4-9} 
\multicolumn{1}{|l|}{}                    &                     &  & 900 Mbps     & 2.19    & 2.53     & 2.37     & 0.06    & 0.34        \\ \hline
\end{tabular}
\end{table*}

\begin{figure*}[h!]
  \begin{subfigure}[t]{.5\textwidth}
    \centering
    \includegraphics[width=0.8\linewidth]{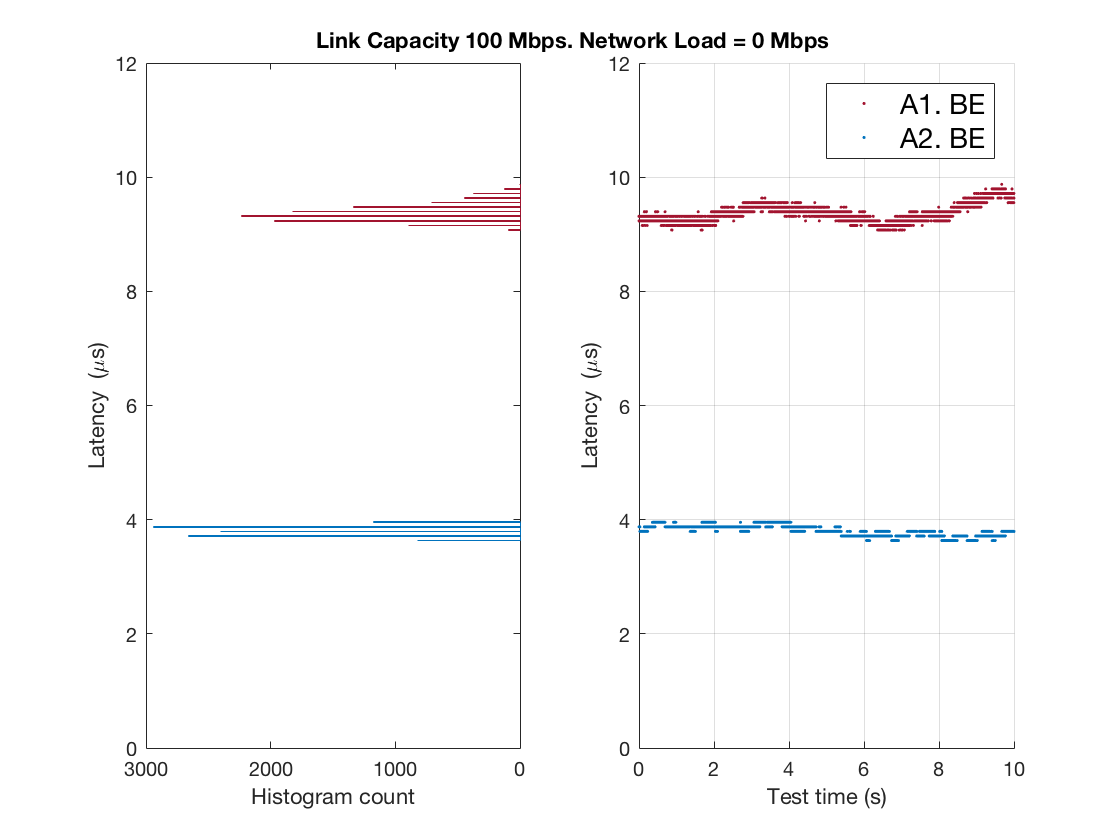}
    \caption{ }
    \label{100mbps_be}
  \end{subfigure}
  \hfill
  \begin{subfigure}[t]{.5\textwidth}
    \centering
    \includegraphics[width=0.8\linewidth]{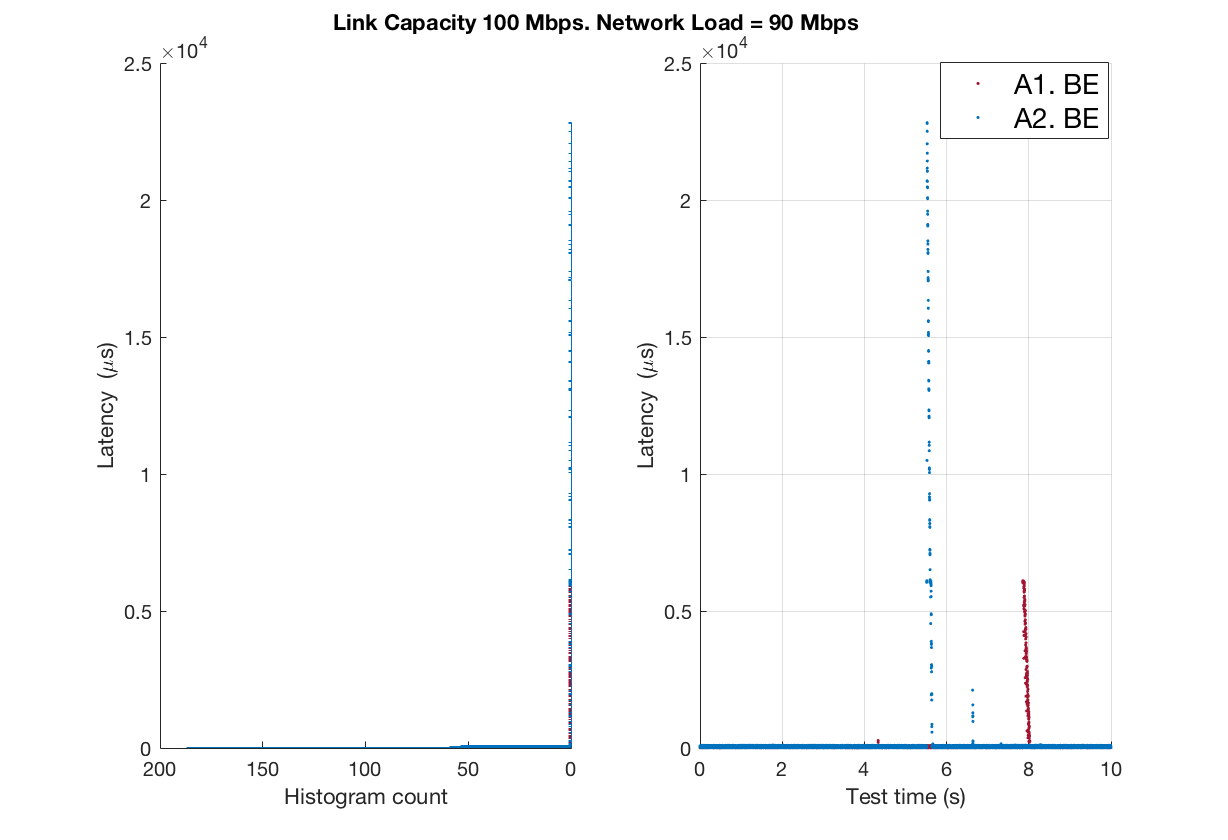}
    \caption{}
    \label{100mbps_be_load}
  \end{subfigure}

  \medskip

  \begin{subfigure}[t]{.5\textwidth}
    \centering
    \includegraphics[width=0.8\linewidth]{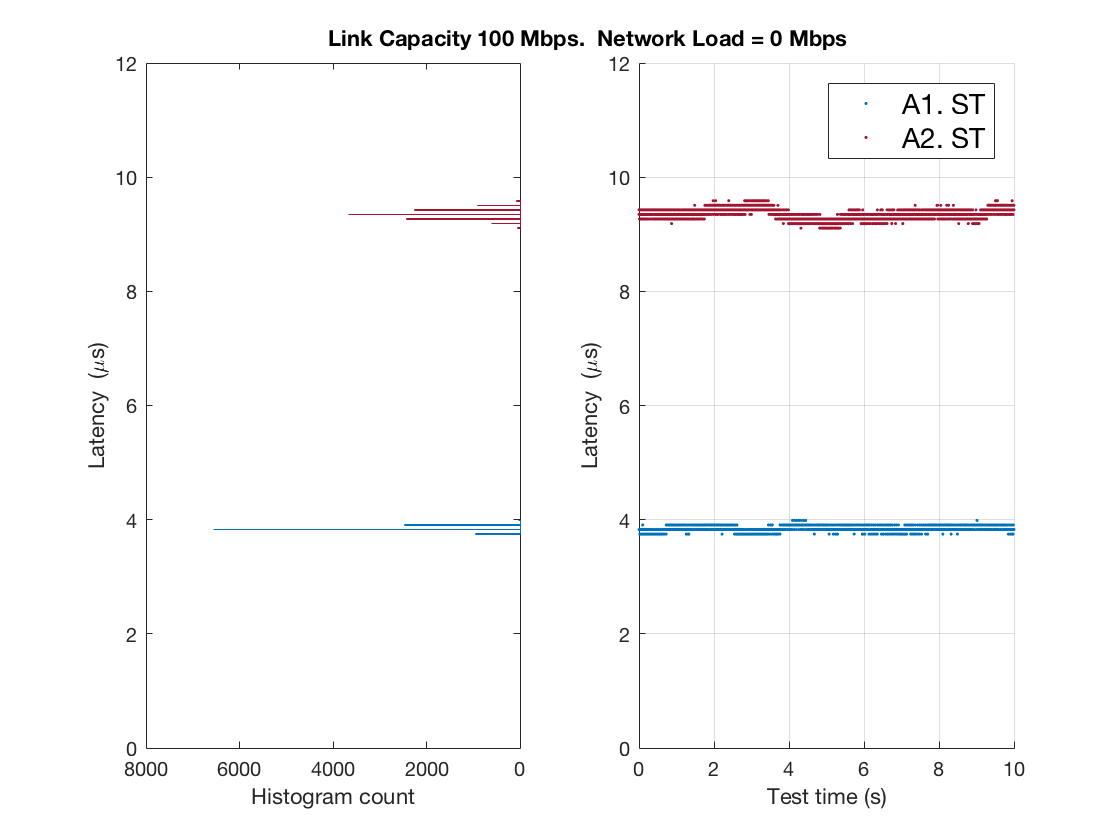}
    \caption{}
    \label{100mbps_st}
  \end{subfigure}
  \hfill
  \begin{subfigure}[t]{.5\textwidth}
    \centering
    \includegraphics[width=0.8\linewidth]{images/100mbps_4.png}
    \caption{}
    \label{100mbps_st_load}
  \end{subfigure}
  
   \medskip

  \begin{subfigure}[t]{.5\textwidth}
    \centering
    \includegraphics[width=0.8\linewidth]{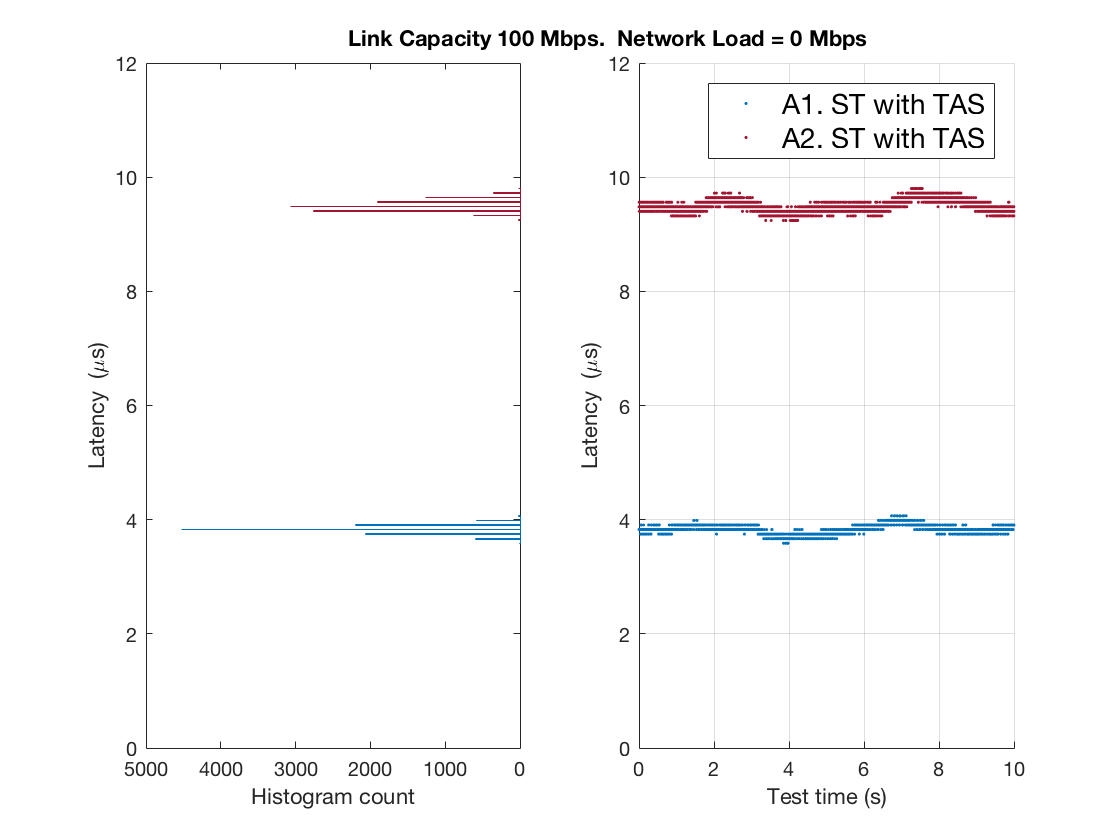}
    \caption{}
    \label{100mbps_st_tas}
  \end{subfigure}
  \hfill
  \begin{subfigure}[t]{.5\textwidth}
    \centering
    \includegraphics[width=0.8\linewidth]{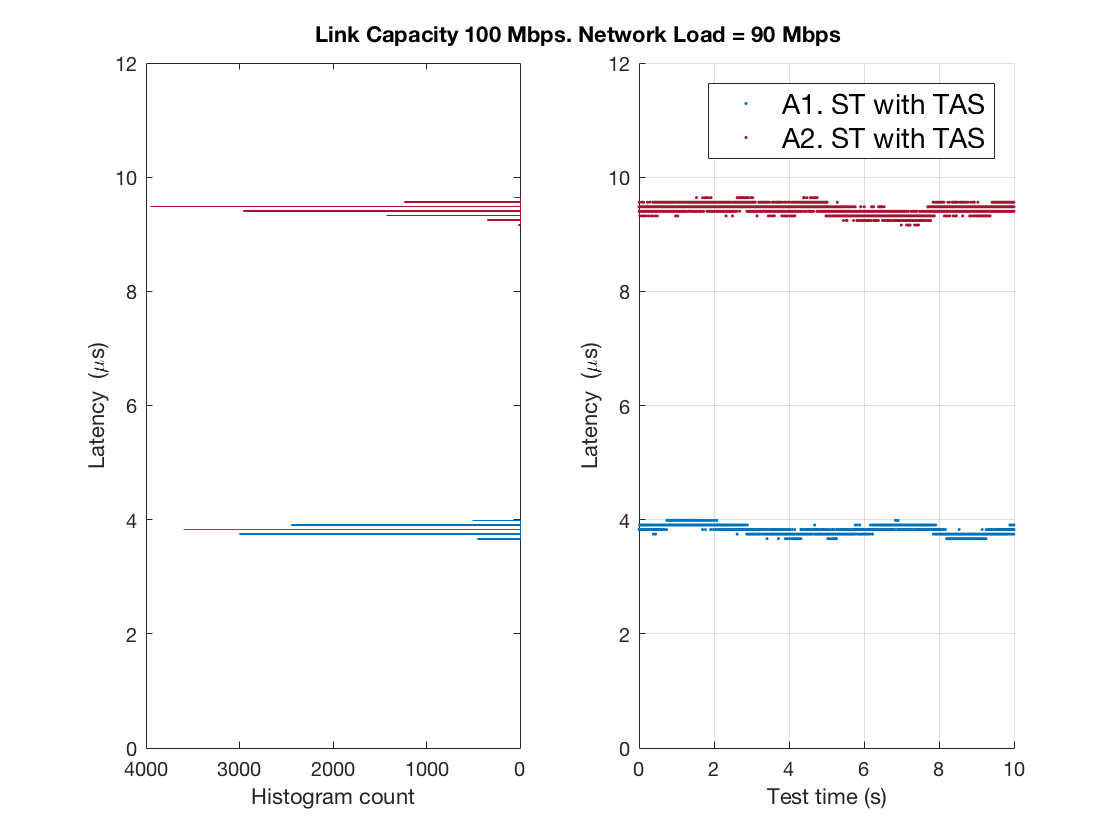}
    \caption{}
    \label{100mbps_st_tas_load}
  \end{subfigure} 
  \caption{\footnotesize Timeplot delay measurements for 10s. Link Capacity 100 Mbps. Terms: \textcolor[rgb]{0,0,1}{A1=Actuator1 (Blue)}, \textcolor[rgb]{1,0,0}{A2=Actuator 2 (Red)}, BE=Best-effort queue, ST=Scheduled Traffic queue.  a) Same priority blocking b)  Same priority blocking with network load c) Lower priority blocking  d) Lower priority blocking with network load e) Using a TAS f) Using a TAS with network load.}
\end{figure*}


\begin{figure*}[h!]
  \begin{subfigure}[t]{.5\textwidth}
    \centering
    \includegraphics[width=0.8\linewidth]{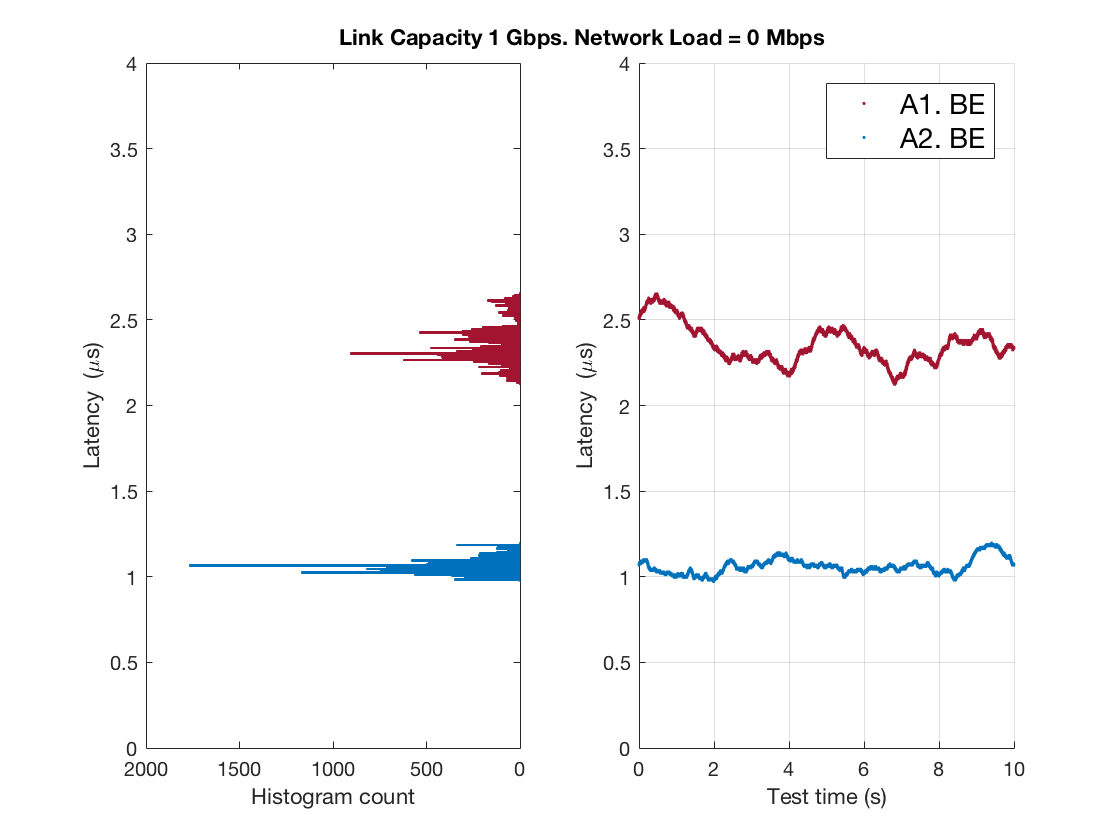}
    \caption{ }
    \label{1gbps_be}
  \end{subfigure}
  \hfill
  \begin{subfigure}[t]{.5\textwidth}
    \centering
    \includegraphics[width=0.8\linewidth]{images/1gbps_2.png}
    \caption{}
    \label{1gbps_be_load}
  \end{subfigure}

  \medskip

  \begin{subfigure}[t]{.5\textwidth}
    \centering
    \includegraphics[width=0.8\linewidth]{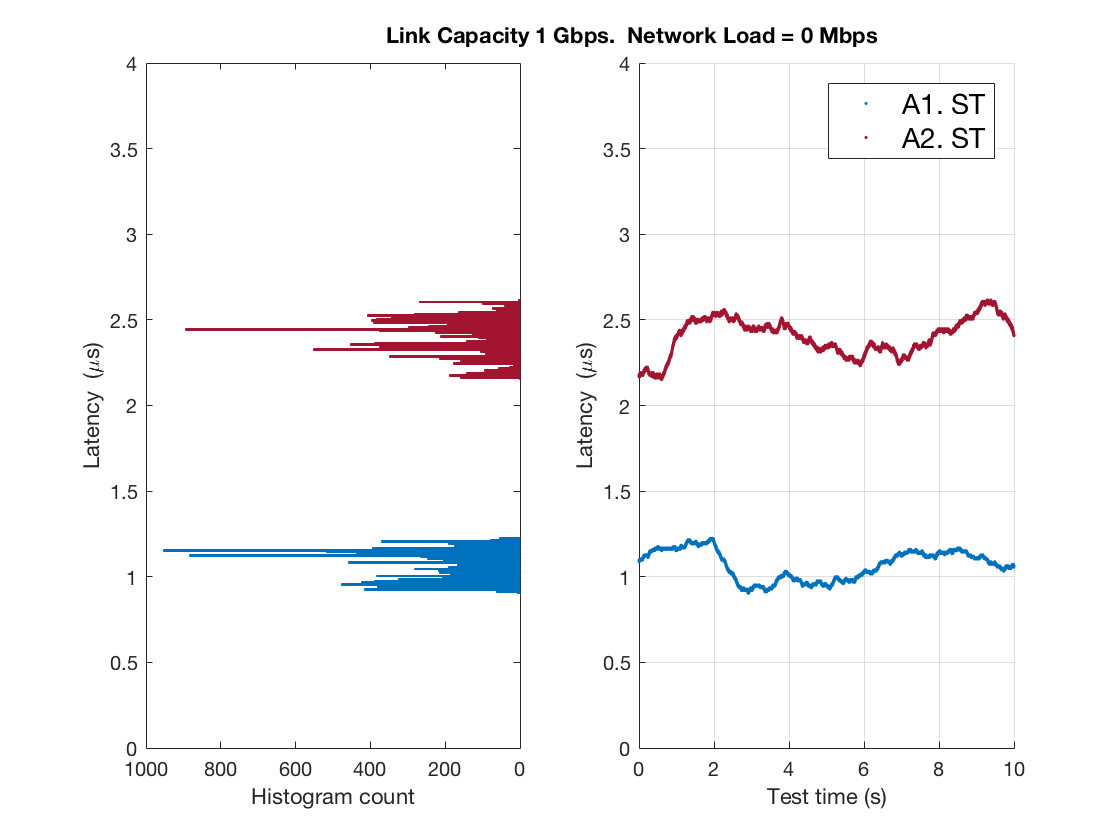}
    \caption{}
    \label{1gbps_st}
  \end{subfigure}
  \hfill
  \begin{subfigure}[t]{.5\textwidth}
    \centering
    \includegraphics[width=0.8\linewidth]{images/1gbps_4.png}
    \caption{}
    \label{1gbps_st_load}
  \end{subfigure}
  
   \medskip

  \begin{subfigure}[t]{.5\textwidth}
    \centering
    \includegraphics[width=0.8\linewidth]{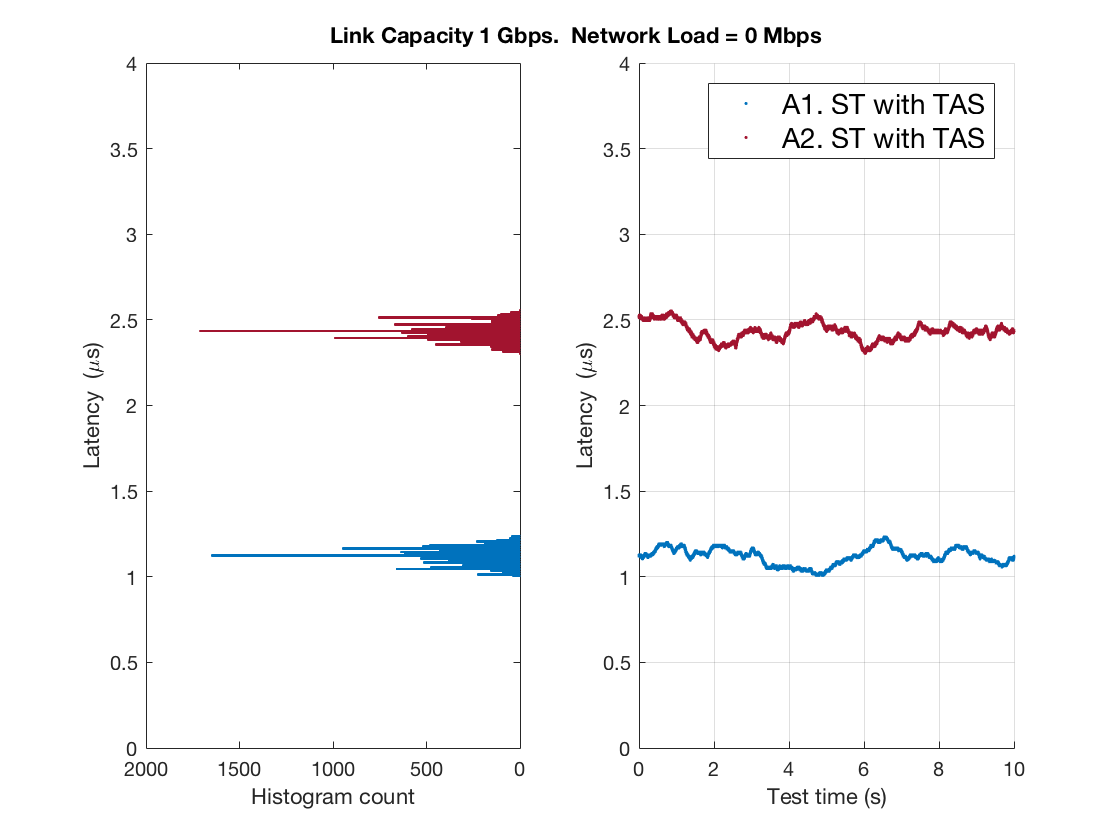}
    \caption{}
    \label{1gbps_st_tas}
  \end{subfigure}
  \hfill
  \begin{subfigure}[t]{.5\textwidth}
    \centering
    \includegraphics[width=0.8\linewidth]{images/1gbps_6.png}
    \caption{}
    \label{1gbps_st_tas_load}
  \end{subfigure} 
  \caption{\footnotesize Timeplot delay measurements for 10s. Link Capacity 1 Gbps. Terms: \textcolor[rgb]{0,0,1}{A1=Actuator1 (Blue)}, \textcolor[rgb]{1,0,0}{A2=Actuator 2 (Red)}, BE=Best-effort queue, ST=Scheduled Traffic queue.  a) Same priority blocking b)  Same priority blocking with network load c) Lower priority blocking  d) Lower priority blocking with network load e) Using a TAS f) Using a TAS with network load.}
\end{figure*}


%





\end{document}